# EXPLORING THE IMPACT OF TRAFFIC SIGNAL CONTROL AND CONNECTED AND AUTOMATED VEHICLES ON INTERSECTIONS SAFETY: A DEEP REINFORCEMENT LEARNING APPROACH

**Amir Hossein Karbasi**
Graduate Research Assistant and Ph.D. Candidate
Department of Civil Engineering
McMaster University, Hamilton, ON, Canada, L8S4L8
Email: Karbaa3@mcmaster.ca

**Hao Yang, Ph.D., Corresponding Author**
Assistant Professor
Department of Civil Engineering
McMaster University, Hamilton, ON, Canada, L8S4L8
Email: yangh149@mcmaster.ca

**Saiedeh Razavi, Ph.D.**
Professor
Department of Civil Engineering
McMaster University, Hamilton, ON, Canada, L8S4L8
Email: razavi@mcmaster.ca

Word Count: 5694 words + 0 table(s) × 250 = 5694 words

**ABSTRACT**
In transportation networks, intersections pose significant risks of collisions due to conflicting movements of vehicles approaching from different directions. To address this issue, various tools can exert influence on traffic safety both directly and indirectly. This study focuses on investigating the impact of adaptive signal control and connected and automated vehicles (CAVs) on intersection safety using a deep reinforcement learning approach. The objective is to assess the individual and combined effects of CAVs and adaptive traffic signal control on traffic safety, considering rear-end and crossing conflicts. The study employs a Deep Q Network (DQN) to regulate traffic signals and driving behaviors of both CAVs and Human Drive Vehicles (HDVs), and uses Time To Collision (TTC) metric to evaluate safety. The findings demonstrate a significant reduction in rear-end and crossing conflicts through the combined implementation of CAVs and DQNs-based traffic signal control. Additionally, the long-term positive effects of CAVs on safety are similar to the short-term effects of combined CAVs and DQNs-based traffic signal control. Overall, the study emphasizes the potential benefits of integrating CAVs and adaptive traffic signal control approaches in order to enhance traffic safety. The findings of this study could provide valuable insights for city officials and transportation authorities in developing effective strategies to improve safety at signalized intersections.

*Keywords*: Deep Q learning; Connected and Automated Vehicles; Traffic Safety; Intersection Safety; Adaptive Signal Control.

# 1. INTRODUCTION

Traffic safety is a paramount concern not, only for the public and private sectors but also for the society at large. The economic impact of traffic crashes in the United States alone amounted to a staggering $340 billion in 2019 (*1*). According to the World Health Organization, traffic crashes resulted in the loss of 1.3 million lives each year (*2*). Therefore, there is an urgent need for research and implementation of diverse and effective solutions to reduce traffic accidents and enhance safety across transportation networks. Among these networks, intersections emerge as particularly perilous, accounting for 36 percent of all traffic accidents as reported by the National Highway Traffic Safety Administration (NHTSA) report (*3*). Additionally, intersections present a significant challenge in terms of waiting time and delays, collectively incurring an annual cost of $8 billion (*4*). Therefore, improving traffic safety and reducing waiting times contribute to the creation of safer and less congested intersections.

In response to the challenges posed by intersections, researchers have explored various solutions, including the optimization of traffic signal management and the integration of innovative vehicle technologies such as Connected and Automated Vehicles (CAVs). Effective traffic signal management plays a pivotal role in alleviating traffic congestion, thereby reducing waiting times and delays (*5, 6*). Furthermore, CAVs hold great promise as a solution to both traffic safety and congestion issues. These vehicles leverage connectivity and automation features to enhance driving behavior and overall performance (*7*). It is anticipated that the adoption of CAVs will lead to improved traffic safety and reduced congestion at intersections. Consequently, combining CAVs with traffic signal control can be a promising approach to enhance traffic safety and alleviate congestion.

In the literature, numerous studies have been conducted to assess the effectiveness of CAVs and traffic signal control in reducing traffic accidents and minimizing waiting time and delays at various intersections layouts (*8–14*). Over the past decade, Deep Reinforcement Learning (DRL) techniques, such as tabular Q learning and Deep Q-Networks (DQNs), have gained wide adoption for traffic signal control at intersections (*8–11*). Additionally, several studies have investigated the impact of CAVs on traffic safety at intersections using microsimulations (*12–14*). Despite the growing body of research on CAVs and DRL to address transportation challenges in safety and congestion, the separate and combined impact of CAVs and DRL-based traffic signal control methods on traffic safety, specifically considering different types of conflicts, has not been yet comprehensively investigated.

To address this research gap, this study aims to answer the following questions:
1. To what extent can DLR- methods based on waiting time reduction can indirectly reduce the number of conflicts at intersections; considering different types of conflicts?
2. How do CAVs contribute to reducing conflicts at intersections, taking into account different types of conflicts?
3. To what extent does the combination of CAVs and DLR traffic signal control methods reduce the number of conflicts at intersections, considering different types of conflicts?

The remainder of the paper is organized as follows. Section 2 reviews existing studies on the impact of CAVs and adaptive traffic signal control on traffic safety and congestion. Section 3 presents the methodology for the study, including the DRL approach to control traffic signals, the car-following models to represent the driving behavior of CAVs and HDVs, the metric for traffic safety assessment, and the definition of different types of conflicts. Section 4 demonstrates the results which are obtained from the simulations. Finally, section 5 summarizes the research findings and provides recommendations and suggestions for future work.

## 2. LITERATURE REVIEW

Numerous studies have been conducted to address the waiting time and delay issues, and safety concerns at intersections (*10, 15*). There are some well-known Traditional traffic signal control methods such as Fixed Time, Self-organizing traffic lights (SOTL) (*16*), SCOOT (*17*), SCATS (*18*), and max pressure (MP) control (*19*) which have been commonly employed in various cities. For instance, the MP method prioritizes the phase with the highest pressure, determined by the difference between the number of vehicles in the entry and exit lanes, when switching the traffic light phases (*20*). Recently, RL-based methods with Q-learning have been widely studied and applied. Prashanth and Bhatnagar (*21*) developed a traffic signal control model using a Q-learning model with function approximation, where the reward was based on the weighted sum of the queue length and elapsed times. This model outperformed previous models, such as fixed time and SOTL. Genders and Razavi (*10*) designed a Deep Q-Network (DQN) utilizing convolutional neural networks (CNNs) and Q-learning methods to control traffic signals at a four-lane intersection. In terms of reducing average cumulative delay, average queue length, and average travel time, this approach outperformed one hidden layer neural network. Fricker and Zhang (*15*) also implemented DQNs at intersections to reduce pedestrian and motorist delays. As in the previous study, these researchers used CNNs and Q learning methods, achieving a significant r reduction in the intersection waiting time.

Various studies have examined the impact of CAVs on traffic safety. For example, The effect of CAVs at intersections was studied by Morando et al. (*22*) using microsimulation. Wiedemann's car-following model was used to demonstrate the behavior of automated vehicles (AVs) and human drive vehicles (HDVs). Their results showed that AVs can reduce traffic conflicts at intersections by up to 65 percent. Arvin et al. (*13*) investigated the impact of AVs and CAVs at intersections and they used Adaptive Cruise Control (ACC) and Cooperative Adaptive Cruise Control (CACC) car-following models to show AVs and CAVs' driving behaviors, respectively. Their results showed that AVs and CAVs can reduce the number of conflicts substantially. In signalized and unsignalized intersections, Karbasi and O'hern (*12*) used the CACC car-following model to assess the effectiveness of CAVs in improving traffic safety. Their results confirmed that CAVs based on the CACC car-following model can reduce conflicts at intersections by up to 100 percent.

Some studies have also investigated the combined impact of Connected Vehicles (CVs), CAVs, and DQNs on traffic safety control, emission, fuel consumption, and waiting time. For instance, Shi et al. (*23*) developed a novel DLR-based model to control traffic signals and used CVs to investigate the effect of traffic control and CVs on a typical four-way intersection's performance. Song and Fan (*24*) employed CAVs and a transfer-based DQN to control traffic signals and examined their impact on waiting time, safety, emission, and fuel consumption, finding positive effects.

Despite some studies exploring intersections' performance under traffic signal control methods or CAVs, there has been no study looking at CAVs and DLR-based traffic signal methods on traffic safety, specifically considering different types of collisions, such as rear-ends and crossings. To bridge this gap, the present study examines the direct or indirect impact of CAVs and DQN-based traffic control methods across different types of conflicts. Additionally, the study considers the short-term and long-term effects of CAVs and DRL-based traffic signal control on traffic safety based on different penetration rates of CAVs.

## 3. METHODOLOGY
This section provides an explanation of the DQN-based traffic control, the simulated environment, the driving behavior of CAVs and HDVs, as well as the safety metrics used in the study. Figure 2 shows the framework of this study. Section 3.1 includes some information related to simulation settings which are related to performing simulations such as network description, data generation, and fixed time signal setting 2. Section 3.2 explains the traffic signal control section in Figure 2. It is worth mentioning that information about fixed-time traffic signals is described in section 3.1. Section 3.3 demonstrates the longitudinal driving behavior in Figure 2. Section 3.4 describes safety metrics that are related to evaluating different scenarios' results based on time-to-collision metrics in Figure 2. Section 4 demonstrates the results of the simulation which is based on determining the impact of DRL-based traffic signal control and CAVs on safety at an intersection in Figure 2.

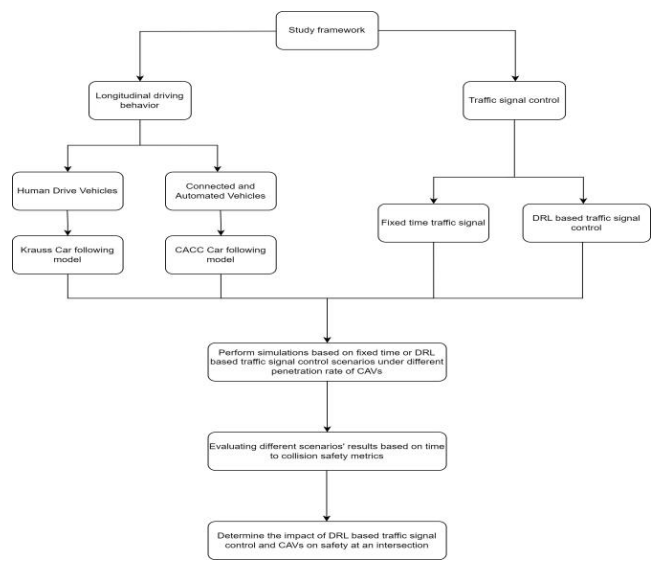

**FIGURE 1 Study framework**

### 3.1 Simulation setting and network description
To simulate the behavior of CAVs, HDVs, and the DQN-based traffic control method, the SUMO simulation software was employed (*25*). Figure 2 demonstrates the network which is one four-way intersection, where the speed limits of all roads are set as 50 km/hr. Each approach in the network comprises four lanes: two lanes dedicated to straight driving, one lane allowing for straight or right-turn movements, and one lane dedicated to left turns. The setting for the fixed time traffic signal is based on the SUMO default traffic signal plan in which the cycle length is equal to 90s. The demand in this network is based on random trip generation using the SUMO built-in feature, called randomTrips.py (*26*). This built-in feature allows users to generate vehicles in each lane randomly. In this paper, 1200 vehicles are generated in the network during 1.5 hours of simulation time. By introducing random trips, the network becomes more flexible and homogeneous when compared to a fixed-route demand model (*12*). The intersection is simulated for 1.5 hours to model the behaviors of individual vehicles and their safety performance.

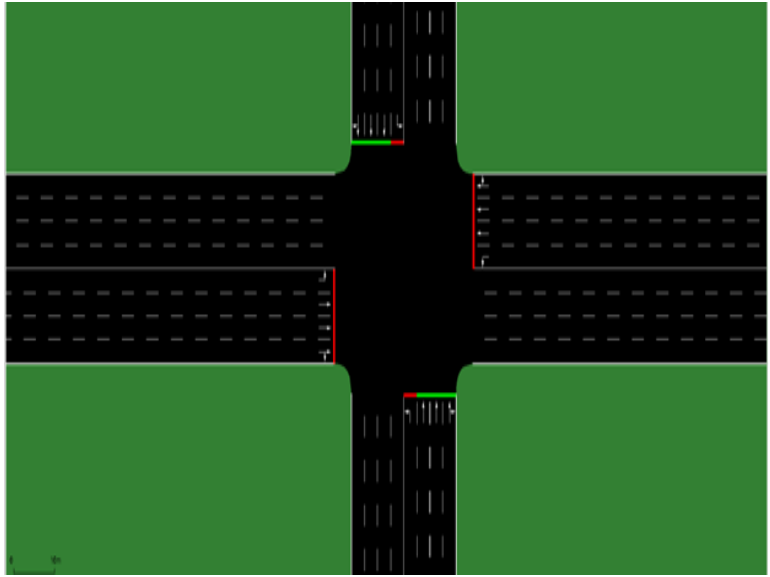

**FIGURE 2 Geometry of one 4-way intersection**

## 3.2 DQN-based traffic signal control system

In this subsection, a DQN-based traffic control system, proposed in (*27*), is employed and advanced to achieve the goal of minimizing vehicle waiting time and travel time delay at signalized intersections. Figure 3 shows the flowchart of the system for controlling traffic signals. SUMO will model traffic conditions at signalized intersections. The agent is a traffic signal controller, which learns to maximize reward based on the deep Q learning method. Once the agent is well-trained, it is anticipated to output optimal signal timing plans for SUMO to minimize delays. The rest of this subsection will present the details of each component in Figure 3.

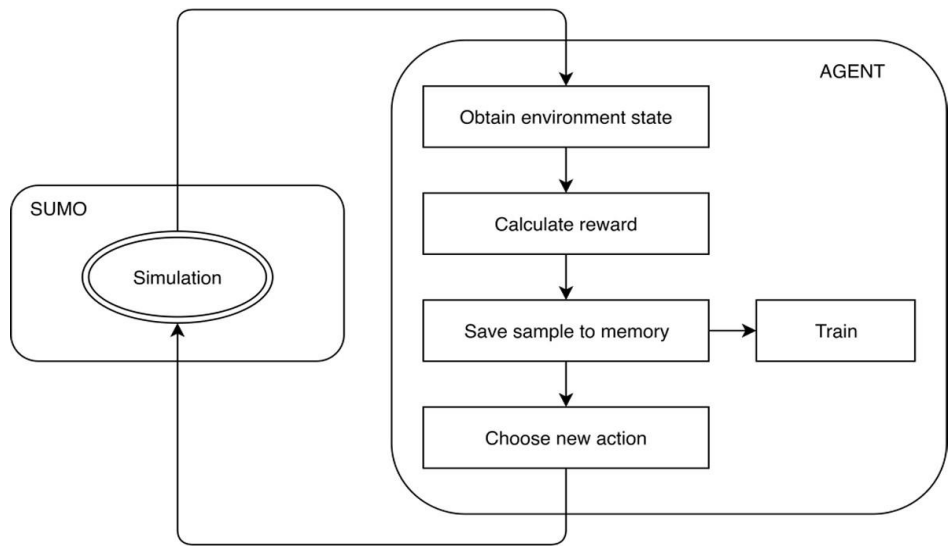

**FIGURE 3 Flow chart of the DQN-based signal control system (*27*).**

*3.2.1 State*

In the context of DQN, the state represents the information of the environment. For this study, the state definition is inspired by Discrete Traffic Signal Encoding (DTSE), which divides a lane of intersection into different cells, each containing data on vehicle position and signal phases (*6*). However, in this study, each cell exclusively gathers vehicles' position information. Each leg of the intersection contains 20 cells in this study based on ref (*27*) study. Each leg of the intersection consists of four lanes, with 10 cells in the left lane and 10 cells in the other three lanes. Therefore, each leg has 20 cells, and the four legs of the intersection have 80 cells. The cell sizes vary depending on their locations; cells near the stop line are smaller (length of vehicle + 2 meters), whereas those farther away are larger. The size of cells has been shown in Figure 4 in meters. For more information about cells see (*27*). It is worth mentioning that CAV's connectivity feature is vehicle-to-vehicle and not vehicle-to-infrastructure and DRL-based traffic signal control doesn't use CAV information as input to its agent.

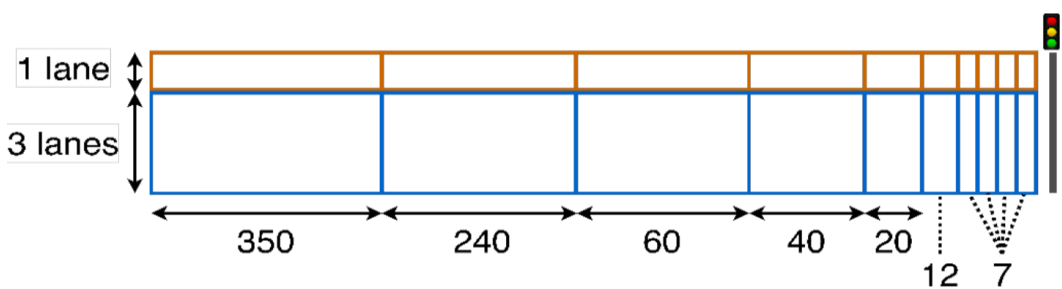

**FIGURE 4 The state representation. The numbers at the bottom show cell size (*27*).**

*3.2.2 Action*

In this study, the DQN agent represents the traffic signal and is capable of selecting from four distinct actions based on its environment. These actions correspond to the specific movements of vehicles at the intersection and determine the activation of the corresponding green phase as follows:

1. North-South Advance (NSA): This action activates the green phase for vehicles traveling straight or turning right in the north and south directions.
2. North-South Left Advance (NSLA): This action triggers the green phase for vehicles making left turns in the north and south directions.
3. East-West Advance (EWA): This action initiates the green phase for vehicles traveling straight or turning right in the east and west directions.
4. East-West Left Advance (EWLA): This action activates the green phase for vehicles making left turns in the east and west directions.

To ensure smooth traffic flow and safety, the duration of the yellow time is set to 4 seconds, serving as a transition period between the green and red phases. Additionally, the green time, representing the duration of the active green phase, is set to 10 seconds and can be incremented by 10s each time if the agent decides to take that action. Through the utilization of these specific actions and the specified durations, the DQN-based traffic signal control system can optimize the traffic flow and minimize conflicts at the intersection.

*3.2.3 Reward function*
The reward function in the DQN is crucial for evaluating the performance of the agent (*28*). Various metrics can be employed as components of the reward function, including the number of vehicles, delays, and waiting times. In this paper, the reward function is designed based on the accumulated total waiting time. Equation 1 illustrates the calculation of the accumulated total waiting time. The variables used in the equation are defined as follows: $awt(i, t)$ represents the time (in seconds) at which vehicle $i$ traverses at a speed of less than 0.1 m/s during time step $t$, and $atwt(t)$ represents the accumulated total waiting time at time step $t$ (*27*). And, $N(t)$ is the total number of vehicles at the intersection at time $t$.

$$atwt(t) = \sum_{i=1}^{N(t)} awt(i, t) \tag{1}$$

With the accumulated total waiting time, the reward function is defined in Equation 2. Here, $r_t$ is the reward at time step $t$ based on the accumulated total waiting time at time $r$ and $t - 1$. The reward function provides feedback to the agent, indicating the effectiveness of its actions. A positive $r_t$ signifies that a beneficial action has been taken by the agent, leading to a reduction in waiting time. Conversely, a negative $r_t$ indicates that the agent's action has been unfavorable, resulting in an increase in waiting time.

$$r_t = atwt(t) - atwt(t - 1) \tag{2}$$

*3.2.4 Deep Q-Learning*
This study utilizes a DQN-based method to control the traffic signal, building upon the concept of Q-learning. Q-learning is an off-policy, model-free RL that aims to select the next best action for a given state. The Q-learning method updates Q values based on the Bellman equation, as defined in Equation (3).

$$Q(s_t, a_t) = Q(s_t, a_t) + \alpha(r_{t+1} + \gamma.maxQ(s_{t+1}, a_t) - Q(s_t, a_t)) \tag{3}$$

where $\{Q(s_t, a_t)\}$ represents the Q-value for state $s_t$ and action $a_t$ at time step $t$. The Q-value is updated using a learning rate $\alpha$ to incorporate the reward $r_{t+1}$ obtained from the time step $t + 1$ and the maximum Q-value for the next state $s_{t+1}$ with discount factor $\gamma$. The term $\gamma$ can vary between 0 and 1 representing the importance of future reward against immediate reward.

In this study, $\alpha$ and $\gamma$ are set as 0.0001 and 0.9 respectively (*28, 29*). Each Q-learning method uses a policy for selecting the action and this paper utilizes the $\varepsilon$-greedy exploration policy. The objective of the $\varepsilon$-greedy is to strike a balance between exploration and exploitation.

Given the vast state space of reinforcement learning in this problem, exploring and recording every state-action combination poses a challenge. To address this challenge, a deep neural network is employed to approximate the Q-learning function. In this setting, the deep neural network approximates the Q-values and the agent selects the action corresponding to the highest Q-value, as illustrated in Figure 5. The deep learning architecture in this study consists of an input layer, representing the states in the network, with 80 cells in total. Each path consists of 20 cells, resulting in 80 cells to represent the entire intersection. The architecture includes 5 hidden layers, each containing 400 neurons, and an output layer with 4 neurons representing Q-values for 4 possible actions. This model showed an acceptable performance in (*27*) study.
	To facilitate the learning process, this study utilizes experience replay (*30*) as the training method. Experience replay involves presenting randomized samples to the agent to provide the necessary information for learning (*27*). The training batch is chosen from a data structure referred to as memory, which is responsible for storing all of the samples collected during the training phase (*27*). Each sample in the memory contains information about the current state, the action taken, the reward received, and the subsequent state.

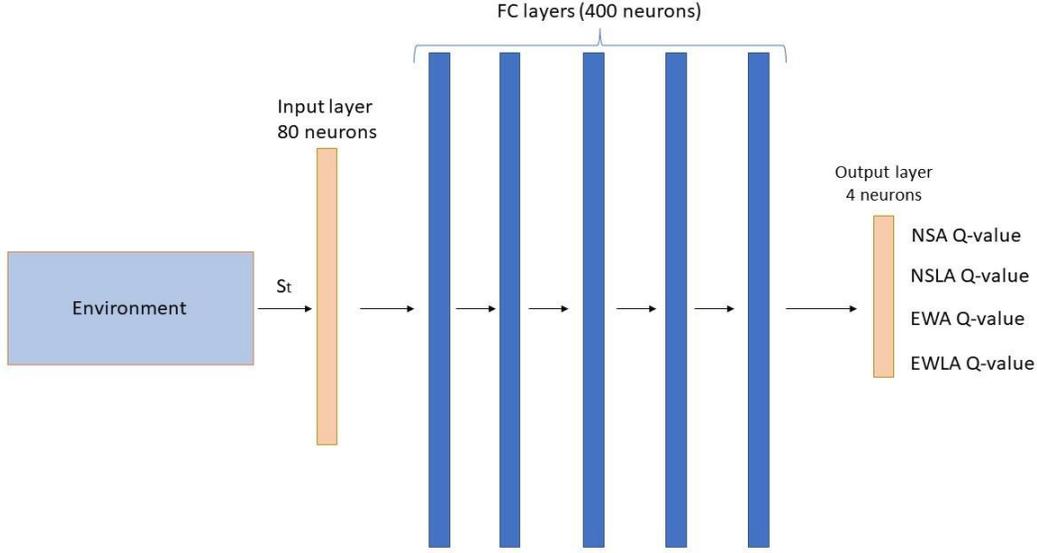

**FIGURE 5 Architecture of the Deep Q Learning model**

### 3.3 CAVs and HDVs driving behaviors
This section describes the driving behavior of CAVs and HDVs in mixed traffic. This study uses car-following and lane-changing models to capture the CAVs and HDVs driving behaviors. In this study, the driving behavior of HDVs is based on the Krauss car-following model (*31*) which is based on safe driving. Equations 4-6 show the speed control in the Krauss car-following model. In equation 4, $v_l(t)$ is the speed of the leading vehicle $i$ in time $t$, $V_{safe}$ represents the safe speed, $t_r$ is reaction time, $g(t)$ is the gap in time $t$ for the leading vehicle, and $b$ is the vehicle maximum deceleration (m/s2). In Equation 4, $V_{safe}$ may exceed the acceleration limit of the vehicle until the next step or may exceed the network speed limit (*12*). Thus, to prevent previous situations, this model presents $V_{des}$ (desired speed) that is shown in Equation 5. The value of $V_{des}$ is based on the minimum value of $V_{max}$, $V_{safe}$, and $v + at$ which is a speed with acceleration capabilities. Moreover, this model accounts for driver imperfection by introducing a random error $\varepsilon$. Equation (6) shows the speed calculation based on random error $\varepsilon$ (*12*).

$$V_{safe} = v_l(t) + ((g(t) - v_l(t)t_r)/((v_l(t) + v_f(t))/2b + t_r)) \qquad (4)$$

$$v_{des} = \min\{v_m, v + acc \cdot t, v_{safe}\} \qquad (5)$$

$$v = \max\{0, \text{rand}[v_{\text{des}} - \varepsilon, v_{\text{des}}]\} \tag{6}$$

CAVs benefit from CACC systems which provide driving assistance and connectivity features. To capture the behavior of CAVs at intersections, this study uses the CACC car-following model (*32–35*), which operates based on four control modes in the SUMO simulator.

The first mode is the speed control mode, where the acceleration of CACC-equipped vehicles is set based on their desired speed, as shown in Equation 7. If the time gap between the vehicle and the leading vehicle is larger than 2 seconds, this mode is activated (*32*).

$$a_{i,t+1} = k_4(v_d - v_{i,t}) \tag{7}$$

In Equation 7, $a_{i,t+1}$ shows the acceleration of vehicle $i$ at the time $t+1$, $v_d$ represents desired speed, $v_{i,t}$ shows the speed of the vehicle in the current vehicle $i$ at the time $t$, $k_4$ is speed gain control, and it is set as $0.4\ s^{-1}$ (*36*). The second mode is the gap control mode, activated under two conditions: when the time gap is smaller than the minimum threshold of 2 seconds or when the distance gap is smaller than 0.2 meters and the speed deviation is smaller than 0.1 m/s (*36*). in this car-following model, first-order transfer functions show the speed of vehicles that are equipped with CACC in the next time step $t+1$.

$$v_{i,k+1} = v_{i,k} + k_5 e_{i,k} + k_6 \dot{e}_{i,k} \tag{8}$$

$$e_{i,k} = v_{i-1,k} - v_{i,k} - t_d \cdot \alpha_{i,k} \tag{9}$$

In Equation 8, $\dot{e}_{i,t}$ shows the first derivative of the gap error $e_{i,t}$. Equation 9 represents gap error calculation. In Equation 9, $t_d$ is the desired time gap for the CACC controller and $\alpha_{i,t}$ is the acceleration at time $t$. In this study, $k_5$ and $k_6$ are $0.45\ s^{-2}$ and $0.0125\ s^{-1}$ respectively (*35*).

The third mode is the Gap-closing control mode, similar to the gap control mode, but with different values of $k_5$ and $k_6$ are $0.005\ s^{-2}$ and $0.05\ s^{-1}$ respectively (*35*). This mode is activated when the time gap is smaller than the minimum threshold. The fourth mode is collision avoidance control mode, in which $k_5$ and $k_6$ are $0.45\ s^{-2}$ and $0.05\ s^{-1}$ respectively (*36*). This mode helps CACC-equipped vehicles to reduce the risk of rear-end collisions.

For both CAVs and HDVs in this study, all parameters except minimum time headway and minimum distance gap are based on default values in the SUMO simulator. CAVs can benefit from smaller time and space headways due to their automation and connectivity (*7*), so the minimum time headway in this study is set to 0.5 seconds for CAVs and 1 second for HDVs. Based on the Atkins (*7*) study, CAVs and HDVs have a minimum distance gap of 0.5 m and 1.5 m, respectively. Moreover, both CAVs and HDVs use SUMO's default lane-changing model (*37*), and all parameters for both types of vehicles are based on the SUMO defaults.

### 3.4 Safety metric
To analyze traffic safety at the signalized intersection, we used a well-known traffic safety metric called time to collision (TTC). TTC represents the time remaining before a collision occurs in a transportation network if both vehicles maintain their current paths and speeds (*38*). In this study, two types of vehicles are considered: CAVs and HDVs, each with different TTC values. Simulations for HDVs record conflicts between vehicles if TTC is equal to or less than 1.5 seconds (*14, 22, 39*). For CAVs, the TTC value is set to 0.5 seconds (*12*). Following Virdi et al.'s (*39*) suggestion, the value of TTC for CAVs is one-third of that for HDVs, resulting in a TTC threshold of 0.5 seconds for CAVs.

To evaluate the safety of the intersections, the Surrogate Safety Measures (SSM) device output in the SUMO simulator is utilized(*40*). The SSM device outputs file provides information about conflicts, including the type of conflict and the corresponding TTC value. Two types of conflicts are used in this study to evaluate the safety situation at intersections:

1. Rear-end conflicts: In this type of conflict, TTC is calculated based on a time period where the following vehicle is faster than the leading vehicle. Equation 10 shows TTC calculation based on rear-end conflicts (*40, 41*):

$$TTC = (((x_{i-1,t} - x_{i,t}) - L_{i-1,t}))/(v_{i,t} - v_{i-1,t}) \tag{10}$$

where $x_{i-1,t}$ is the leader position at the time $t$, and $v_{i-1,t}$ is the leader speed at the time $t$, $x_{i,t}$ follower position at time $t$, $v_{i,t}$ is the speed of the following vehicles at the time $t$ and $L_{i-1,t}$ is the length of vehicle at the time $t$.

2. Crossing conflict: This type of conflict occurs when vehicle A's expected time to exit the conflict area is longer than vehicle B's expected time to enter the conflict area, and vehicle A has a shorter expected time to enter the conflict area (*40*). Equation (11) shows the TTC calculation for crossing conflicts, where $v_B$ is B's current speed and $S_b$ is B's distance to the conflict area (*40*).

$$TTC = S_B/v_B \tag{11}$$

## 4. RESULTS

A microscopic traffic simulation of the intersection in Figure 2 is studied to evaluate the benefit of the proposed DQL system. This section includes two subsections. subsection 4.1 demonstrates the results of DRL-based traffic signal control under six scenarios. In each scenario, the penetration rate of CAVs increases by 20%. Subsection 4.2 shows the impact of CAVs and DRL-based traffic signal control on traffic safety.

### 4.1 DQN-based traffic signal control results

As a means of evaluating the effectiveness of DQN-based traffic signal control, the learning curve is presented in Figure 6 using the cumulative negative reward for each PR, and the cumulative delay for each PR is presented in Figure 7. After 40 episodes, the cumulative negative reward has decreased significantly, demonstrating that DQN-based traffic signal control has learned the policy efficiently. Figure 6 demonstrates that DRL-based traffic signal control reduces negative reward from about 30000 to 5000 for each penetration rate of CAVs. Additionally, the results show that when the CAV penetration rate is 100, the commutative delay decreases in the first episode as compared to 0. However, When the agent is trained after 40 episodes, there is no significant difference in commutative delay when the penetration rate of the CAVs increases. Therefore, once a network's delay has been reduced by a certain amount, changing driving behavior cannot significantly affect it. Since the hardware limitation requires the agent to be trained with 40 episodes, future research will focus on training the agent with more episodes, such as 50 or 100, to improve reward performance. It may also be possible to investigate how the delay changes as the penetration rate of CAVs increases after 100 episodes.

Based on the results presented in Figure 7, it is evident that DQN-based traffic signal control effectively reduced cumulative delays for all PRs after 40 episodes, demonstrating that this control method can effectively manage traffic at intersections. Moreover, Figure 7 shows that DRL-based traffic signal control can reduce delay substantially up to about 10000 s in each penetration rate of CAVs.

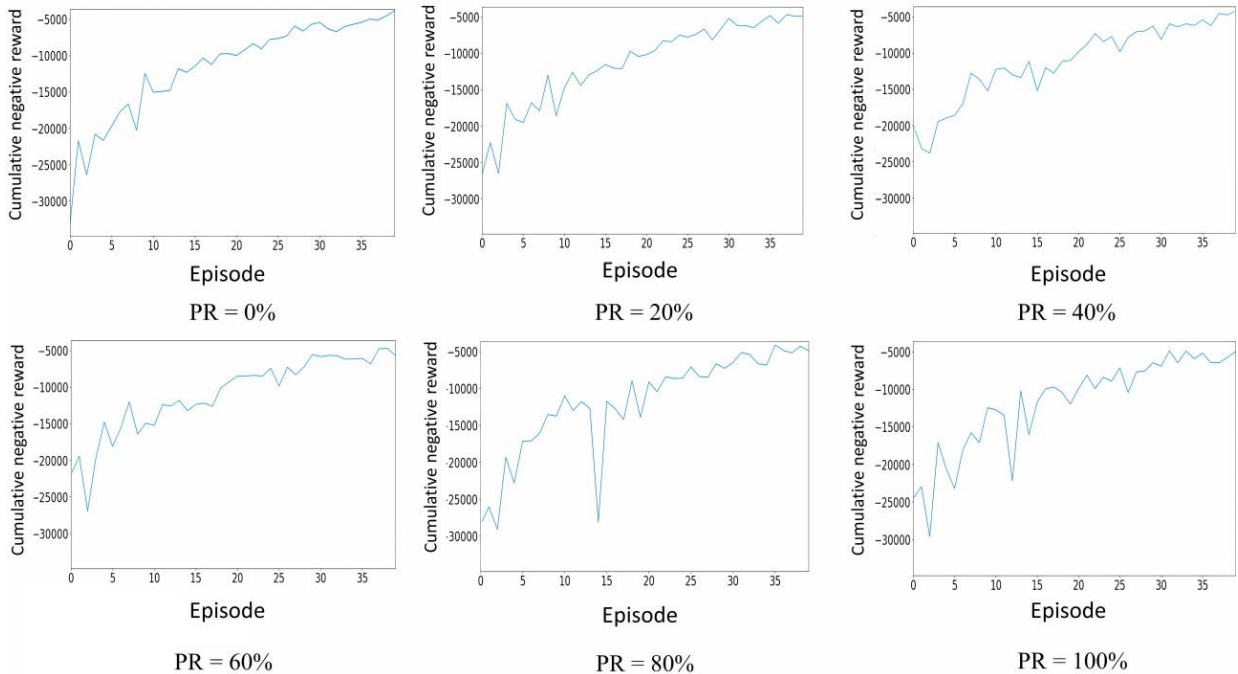

**FIGURE 6 Cumulative negative rewards at different PRs**

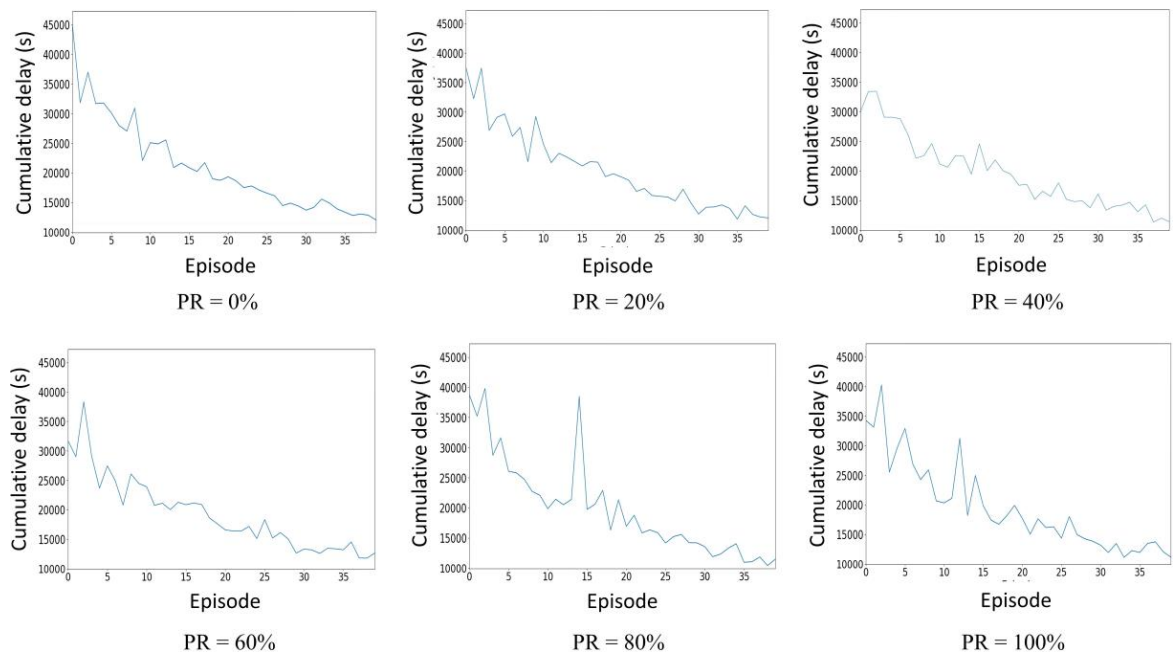

**FIGURE 7 Cumulative delays at different PRs**

## 4.2 Impact of CAVs and DQN-based traffic control on intersection safety

This study examines how the combination of CAVs and DQN-based traffic signal control may affect the number of conflicts at intersections based on different types of conflicts. The results are based on the total number of conflicts, the number of rear-end conflicts, and the number of crossing conflict results. Figure 8 demonstrates the impact of CAVs and DQN-based traffic signal control on traffic safety. Generally based on Figure 8 information, at PR 0%, the Fixed time-based traffic signal shows 2835 conflicts, higher than the based traffic signal which shows 817 conflicts. By PR 20%, the DQN-based traffic signal showcases its efficiency with 629 conflicts, whereas the Fixed time traffic signal records 2287 conflicts. At PR 40%, the gap is more pronounced: the DQN-based traffic signal reports 493 conflicts as opposed to 1766 conflicts of the Fixed time system. And, echoing the trend, by PR 100%, the DQN-based traffic signal eliminates all conflicts, while the Fixed time system struggles with 98 conflicts. In detail, Three different control strategies are compared in the figure:

1. **DQN-based traffic signal control only**: When the PR of CAVs is 0, Figure 8 compares the total numbers of conflicts in fixed time and DQN-based traffic signal control without CAVs. With the DQN-based control, the number is reduced from 2835 to 817, i.e., a 71% reduction. This verifies the safety advantage of the DQN-based control.
2. **CAVs Only Without DQN-Based Control**: The blue line in Figure 8 illustrates the impact of Connected and Autonomous Vehicles (CAVs) on traffic safety, independent of DQN-based traffic signal control. This graph demonstrates that safety increases with the CAV penetration rate (PR). Specifically, at a PR of 100%, conflicts decrease to 98—a reduction of more than 96% compared to when the PR is 0%.
3. **Combined CAVs and DQN-Based Traffic Signal Control**: The red line in Figure 8 depicts the combined effect of CAVs and DQN-based traffic signal control on safety across various CAV PR levels. The results highlight that with 100% CAV PR and DQN-based control, the number of conflicts drops to zero, indicating that the integration of these technologies can potentially create a collision-free intersection.

These results are important from a time period's point of view. For instance, when CAVs with a PR of 80% operate under fixed-time traffic signal control, they experience around 650 conflicts. On the other hand, when CAVs with a PR of 20 percent are under DQN-based traffic signal control, they experience around 630 conflicts. This result is crucial since it's expected that 100 percent of drivers to operate CAVs by 2050 (*42*) whereas only 20 percent of passenger vehicles will be highly automated by 2030 (*43*). Thus, this result indicates that the anticipated 20 percent CAVs with DQN-based traffic signal control in 2030 will have a similar impact on safety compared to that of CAVs in fixed-time traffic signals with 100 percent penetration, which is anticipated for the year 2050 and beyond.

Moreover, Figure 9 shows a more detailed analysis of rear-end conflicts and crossing conflicts in both fixed-time and DQN-based traffic signal control. Generally, Comparing rear-end conflicts between the fixed-time traffic signals and the DQN-based traffic signal control, a noticeable distinction is seen across different PR levels. At PR 0%, the DQN-based traffic signal control system shows 592 conflicts, remarkably lower than the 2582 conflicts of the fixed-based traffic signal control. At PR 20%, this gap is maintained with the DQN-based traffic signal control reporting 477 conflicts against the fixed-based traffic signal control system's 2109 conflicts. By PR 40%, DQN-based traffic signal control further reduces its conflicts to 380, while the fixed time-based traffic signal system has 1629 conflicts. Significantly, by PR 100%, the DQN-based traffic sig-nal control system entirely nullifies rear-end conflicts, whereas the Fixed time-based traffic signal control system shows 98 conflicts. For crossing conflicts, both systems have a closer starting point at PR 0% with the fixed time-based system at 253 conflicts and the DQN-based system slightly lower at 225 conflicts. However, as PR increases, the gap widens. By PR 20%, the DQN-based system demonstrates 152 conflicts, outdoing the fixed time-based system's 178 conflicts. At PR 40%, the DQN-based system further reduces its tally to 113 conflicts, contrasting with the fixed time-based system of 137 conflicts. Remarkably, at PR 100%, the DQN-based system and the Fixed time-based system reduce the number of conflicts to 0. Similar to Figure 8, three different control strategies are compared in detail:

1. **DQN-based traffic signal control only**: At PR=0, Figure 9 illustrates that the rear end and crossing number of conflicts in DQN-based traffic signal control are 592 and 225, respectively, and 2582 and 253 under the fixed-time signal. These results show that DQN-based traffic signal control reduces 80% of the rear-end conflicts compared to fixed traffic signals which means that DQN-based traffic signal control is very effective in improving the safety of intersections from the rear-end conflicts point of view substantially. However, the results show that the impact of DQN-based traffic signal control is not very effective in improving the safety of intersections from crossing conflicts. A possible reason is that DQN-based traffic signal control can not have much effect on vehicles that pass the intersection.
2. **CAVs only without DQN-based control**: The blue line in Figure 9 shows the safety effect of CAVs without DQN-based traffic signal control. The results demonstrate that as the penetration rate of CAV increases the number of rear-end and crossing conflicts decreases. When the penetration rate of CAVs is 0 and 100, the number of rear-end conflicts are 2582 and 98 respectively which shows that CAVs decrease the number of conflicts dramatically. Moreover, the results show that CAVs decrease the number of crossing conflicts to 0 which is a substantial effect on safety.
3. **Combined CAVs and DQN-based traffic signal control**: The red line in Figure 9 demonstrates the combined impact of CAVs and DQN-based traffic signal control on traffic safety. The results demonstrate that when the penetration rate of CAVs is 100 and the traffic signal is based on DQN-based traffic signal control, the number of rear-end and crossing conflicts reduces to 0 which demonstrates that the combination of CAVs and DQN-based traffic signal control can create a collision-free intersection based on rear end and crossing conflicts.

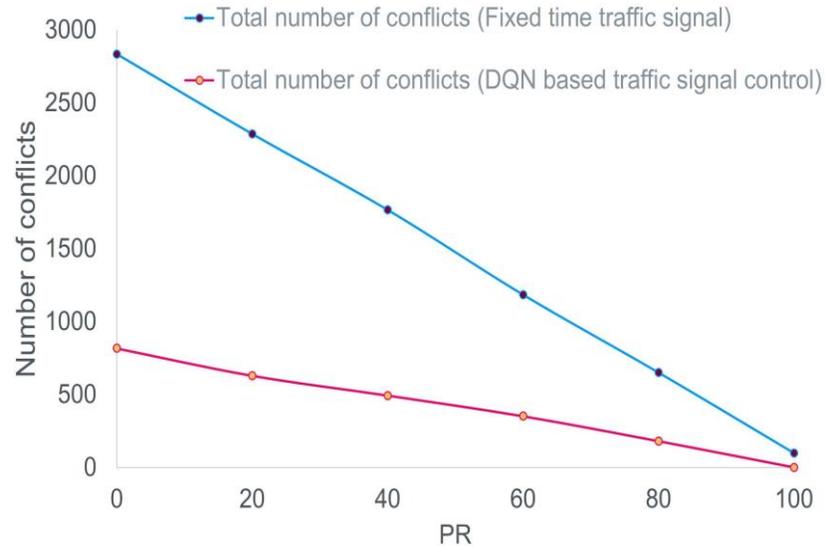

**FIGURE 8 Change of a total number of conflicts based on fixed time traffic signal and DQN-based traffic signal control.**

Moreover, the results show that although DQN-based traffic signal control can not reduce crossing conflict effectively, CAVs can reduce crossing conflict dramatically which means that CAVs can compensate inability of DQN-based traffic signal control in reducing crossing conflicts.

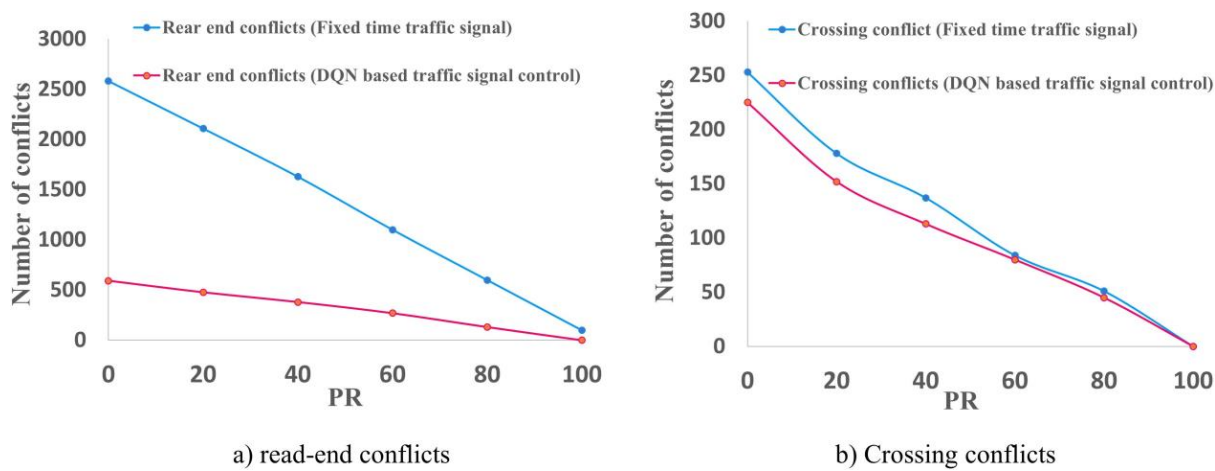

a) read-end conflicts

b) Crossing conflicts

**FIGURE 9 Change of total number of conflicts based on rear-end and crossing conflicts.**

## 5. CONCLUSION

This study investigated the impact of CAVs and a DQN-based traffic signal control approach on traffic safety at signalized intersections. The DQN-based traffic signal control model was used to optimize traffic signal timings to minimize the cumulative waiting time. To model mixed traffic, the study utilized the CACC model for CAVs and the Krauss car-following model for HDVs. Two types of conflicts, rear-end conflicts and crossing conflicts, were evaluated with TTC to measure traffic safety. The system evaluation provided valuable insights into the effectiveness of integrating CAVs with advanced traffic signal control methods to achieve optimal traffic safety and reduce conflicts at intersections. The results clearly demonstrated that both CAVs and DQN-based traffic signal control could independently contribute to improving safety at intersections. Moreover, when combined, the effect of CAVs and DQN-based traffic signal control is even more significant toward achieving collision-free intersections.

In addition, the study found that the effect of CAVs and DQN-based traffic signal control remained similar regardless of the penetration rate (PR) of CAVs in fixed-time traffic signal scenarios. This suggested that in the short term, the impact of CAVs and DQN-based traffic signal control was comparable to the long-term effects of CAVs only on traffic safety. As a result, poli-cymakers could consider the combination of these technologies as a short-term solution to address traffic safety issues effectively. By benefiting from these advancements, cities and communities can move closer to achieving safer and more sustainable transportation systems in the future.

In the future, traffic signal control can be further enhanced by exploring alternative types of DQNs, such as Double Deep Q Networks (DDQN). Investigating the efficacy of DDQNs and other advanced reinforcement learning techniques can lead to further improvements in optimizing traffic signal timings and traffic flow management. The study didn't validate the methodology across different traffic volumes or compare delays with established methods like Webster's. Key sustainability metrics, such as fuel consumption, were overlooked. Hence, researchers should explore the effects of CAVs and traffic signal control across a spectrum of traffic demands, from low to high. This examination should also encompass the environmental implications of both CAVs and DQN-based traffic signal management. Understanding how CAVs and traffic signal control perform under different traffic conditions can provide valuable insights for real-world implementation. Furthermore, future studies can focus on exploring the collective impact of vehicle trajectory optimization, traffic signal control, and CAVs on traffic safety. Analyzing these factors together can lead to comprehensive strategies for improving overall traffic safety and reducing the frequency of accidents at intersections. To robustly ascertain the system's benefits, it's recommended to expand the research to multiple intersections or a broader network.

**ACKNOWLEDGEMENTS**
All the authors have contributed to the model development, experiment design, data analysis, and paper preparation and review.